\documentclass[letterpaper, 10 pt, conference]{ieeeconf}  

\IEEEoverridecommandlockouts                              
\usepackage{graphicx} 
\usepackage{epsfig} 
\usepackage{amsmath} 
\usepackage{amssymb} 
\usepackage{pifont}
\usepackage{soul,color} 
\usepackage{mathtools}
\usepackage{booktabs}
\usepackage{multirow}
\usepackage{enumerate}
\usepackage{balance}
\usepackage{gensymb}
\usepackage{tabulary}
\usepackage[numbers,sort&compress]{natbib}
\usepackage{dblfloatfix}
\usepackage{layouts}
\usepackage{algorithm}
\usepackage{algpseudocode}
\usepackage{lipsum}
\usepackage{subcaption}
\usepackage{siunitx}
\usepackage{hyperref}
\usepackage[font=scriptsize,labelfont=bf]{caption}

\let\labelindent\relax
\usepackage{enumitem}

\graphicspath{{figures/}}

\newcommand{\figGap}[0]{\vspace{-1.3\baselineskip}}

\newcommand{\argmin}{\operatorname*{argmin}}

\title{\bf 
PatchGraph: In-hand tactile tracking with learned surface normals
\vspace{-2mm}
}
\author{Paloma Sodhi$^{1,2}$, Michael Kaess$^{1}$, Mustafa Mukadam$^{2}$, and Stuart Anderson$^{2}$\\[2mm]
$^{1}$Carnegie Mellon University, $^{2}$Meta AI Research
\thanks{\footnotesize{Code and supplementary material can be found on \url{https://psodhi.github.io/tactile-in-hand}}}
}

\begin{document}

\maketitle
\thispagestyle{empty}
\pagestyle{empty}

\begin{abstract}
We address the problem of tracking 3D object poses from touch during in-hand manipulations. Specifically, we look at tracking small objects using vision-based tactile sensors that provide high-dimensional tactile image measurements at the point of contact. While prior work has relied on a-priori information about the object being localized, we remove this requirement. Our key insight is that an object is composed of several local surface patches, each informative enough to achieve reliable object tracking. Moreover, we can recover the geometry of this local patch online by extracting local surface normal information embedded in each tactile image. We propose a novel two-stage approach. First, we learn a mapping from tactile images to surface normals using an image translation network. Second, we use these surface normals within a factor graph to both reconstruct a local patch map and use it to infer 3D object poses. We demonstrate reliable object tracking for over $100$ contact sequences across unique shapes with four objects in simulation and two objects in the real-world.
\end{abstract}

\section{Introduction}
\label{sec:introduction}

We focus on the problem of tracking 3D object poses during in-hand manipulations using tactile image measurements from vision-based tactile sensors \cite{lambeta2020digit, yuan2017gelsight}. Specifically, we look at tracking small objects without prior geometric models. For instance, a dexterous robot operating in a real-world household environment will need to manipulate novel household objects for which CAD models may not be available. We address the question: Can an object be tracked precisely enough for in-hand manipulation using only local measurements of its geometry?

Prior work has looked at the object tracking problem primarily in the context of planar pushing \cite{sodhi2021tactile, suresh2021shape, lambert2019joint, yu2018realtime}. However, the problem of 3D in-hand manipulation poses additional challenges. Firstly, the motion is less constrained such that different object motions can explain the same measurements. Second, physics priors, such as quasi-static planar pushing models, are less informative in the 3D case. Hence, prior work on in-hand object tracking using tactile feedback has relied on a-priori information about the object being localized, such as 3D global models or a database generated by a simulator \cite{wang2021gelsight, bauza2020tactile, liang2020hand, bauza2019tactile}.

Our key insight is that the tactile object tracking problem can be efficiently decomposed in two ways. First, we can decompose an object into many smaller local surface patches, which can be treated independently. Second, most of the information needed to infer the local surface patch geometry is already embedded in corresponding tactile images.


We find that reliable tracking is achievable with only a \textit{local patch}---a fused map created from a sequence of key frame images within a continuous contact episode. For instance, even though two objects can have very different global geometries, they can contain very similar local patches that suffice for tracking.

To both create the local patch and track motion relative to it, we must fuse multiple tactile image measurements online while inferring the latent object poses. We formulate this as an inference problem over a factor graph that offers a flexible and efficient way to fuse such information while incorporating other priors derived from physical and geometric constraints \cite{dellaert2020factor,dellaert2017factor}.

\begin{figure}[!t]
	\centering
	\includegraphics[width=\columnwidth]{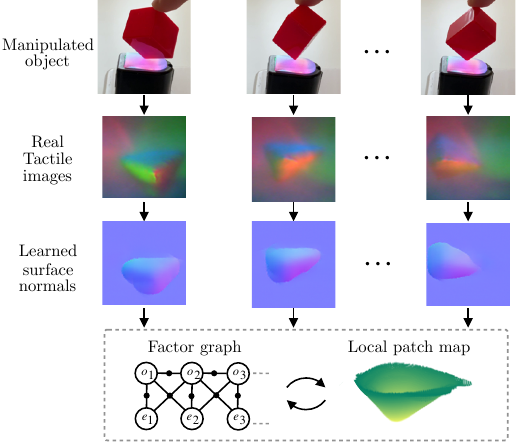}
	\caption{\footnotesize Tracking latent 3D object poses from a tactile image sequence during in-hand manipulation. We solve this as an inference over a factor graph that does not rely on having prior object models.}
	\label{fig:cover}
	\figGap
\end{figure}

What makes for a good representation for a local patch? An idealized tactile image captures the surface normals of the gel's reflective layer, based on the color and intensity of illumination at each pixel. Hence, it is natural to learn a mapping from image to gel surface normals. While objects may have varying global shapes, the learned surface normal mapping can generalize across these different shapes, because the relationship between pixel intensities and surface normals depends only on the sensor configuration and the local contact geometry. Hence, we infer the surface normals at each pixel in the tactile image, then integrate those normals to create a 3D model of the visible section of the local patch.

We propose a novel two-stage approach for tracking objects in-hand without any prior object model information (Fig. \ref{fig:cover}). 
First, we learn a mapping from tactile images to surface normals using an image translation network. Second, we use these surface normals within a factor graph to both reconstruct a local patch map and use it to infer 3D object poses. Our key contributions are:
\begin{enumerate}
    \item A factor graph formulation for 3D in-hand tactile tracking that does not rely on prior object models.
    \item A factor that works across different global object shapes by relying on local patches generated from learned surface normals.
    \item Empirical evaluation on both simulation and real-world trials.
\end{enumerate}

\section{Related Work}
\label{sec:relatedwork}

\textbf{Factor graphs for localization}
Localization and mapping problems are increasingly formulated as optimization objectives that leverage the inherent sparsity of the problem to give tractable and more accurate solutions over filtering approaches \cite{cadena2016past}. Factor graphs are a popular way for solving such optimization objectives \cite{dellaert2020factor, dellaert2017factor, sodhi2021tactile, czarnowski2020deepfactors, hartley2018hybrid}. They offer a flexible way to fuse multiple measurements while being computationally efficient to optimize. Factors in the graph encode local potentials on variables such as observation models between measurements and states as well as other priors such as physics and geometry. We formulate our problem using a factor graph based framework in this paper.

\textbf{Vision-based touch sensing}
The advent of vision-based tactile sensors \cite{lambeta2020digit, yuan2017gelsight, donlon2018gelslim, yamaguchi2016fingervision} has enabled high-dimensional tactile image measurements that capture the local deformation at the point of contact. Recent work has also looked at creating accurate simulation models for such sensors \cite{wang2020tacto, si2021taxim, agarwal2020simulation}. These sensors are being explored for various tactile manipulation applications. One class of approaches use tactile images directly as local feedback to solve for control actions on tasks such as object insertion \cite{dong2021tactilerl}, box packing \cite{dong2019boxpacking}, and in-hand manipulations \cite{she2020cable, lambeta2020digit}. However, such representations tend to overfit to the particular task or require significant amount of data to generalize across tasks. Our work focuses on extracting a state representation like the global object pose that is easy to use and generalizes across different downstream control and planning tasks.

\textbf{Estimation from touch}
Prior work on estimating states from touch during manipulation has included filtering methods \cite{izatt2017tracking, saund2017touch, koval2015mpf, pezzementi2011object}, learning-only methods \cite{sundaralingam2019robust, li2014localization}, methods utilizing prior model information \cite{bauza2020tactile, liang2020hand, bauza2019tactile}, and graph-based optimization for planar pushing \cite{yu2018realtime, lambert2019joint, suresh2021shape, sodhi2021tactile}. In particular, graph-based optimization offers benefits such as being more accurate than filtering, an ability to incorporate analytic as well as learned models, and can be solved in real-time making use of efficient, incremental solvers \cite{kaess2008isam, kaess2012isam2, sodhi2020ics} in literature.

Of these different approaches, the work in \cite{bauza2019tactile, liang2020hand, bauza2020tactile, wang2021gelsight} is most closely related in terms of the final objective of tracking 3D object poses during in-hand manipulations.These, however, require prior object model information either as offline models \cite{wang2021gelsight, bauza2019tactile, liang2020hand} or a database from a simulator \cite{bauza2020tactile}. In contrast, we do not require a prior model of the object being tracked. Instead, we build a local patch map on the fly for the current contact episode and use that within a factor graph framework.

\begin{figure*}[!t]
	\centering
	\includegraphics[width=\textwidth]{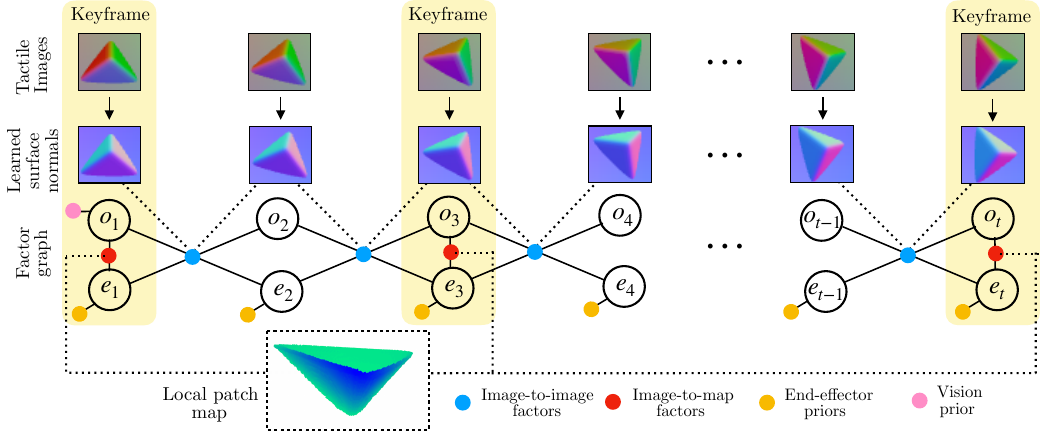}
	\caption{\footnotesize Overview of our factor graph formulation, where variables are object poses and factors are constraints on poses. We first predict learned surface normals from tactile images used to reconstruct a 3D point cloud. We then create a set of factors. Image-to-image factors (blue) 
	encode relative pose between two point clouds. Image-to-patch factors (red) encode relative pose between point cloud and a fused local patch. We also include global pose priors as unary factors on end-effector (yellow) and first object pose (pink).}
	\label{fig:approachOverall}
	\figGap
\end{figure*}

\section{Problem Formulation}
\label{sec:probform}
We begin by formalizing the estimation problem as factor graph optimization. A factor graph is a bipartite graph with two types of nodes: variables $x\in \mathcal{X}$ and factors $\phi(\cdot):\mathcal{X}\rightarrow\mathbb{R}$. Variable nodes are the latent states to be estimated, and factor nodes encode potentials on these variables from observations, physics, or geometry (Fig. \ref{fig:approachOverall}). 

We solve for the maximum a posteriori (MAP) objective $\hat{x}$ by maximizing product of all factor graph potentials, i.e.,
\begin{equation}
	\begin{split}
		\label{eq:eq3.1}
		\hat{x}^{}& = \underset{{x}}{\operatorname{argmax}}\ \prod_{i}^{} \phi_i({x_i}) \\
	\end{split}
\end{equation}

To solve this inference objective efficiently, we assume $\phi_t(x)$ to be Gaussian factors corrupted by zero-mean normally distributed noise. Under Gaussian noise model assumptions, MAP inference is equivalent to a nonlinear least-squares objective \cite{dellaert2017factor}, i.e.,
\begin{equation}
	\begin{split}
		\label{eq:eq3.2}
		&\phi_i({x_i})\propto \exp\left\{-\frac{1}{2}||f_i(x_i;z_i)||_{\Sigma_i}^2\right\} \\
		\Rightarrow \ & \hat{x}=\underset{{x}}{\operatorname{argmin}}\ \frac{1}{2}\sum_{i}^{}||f_i(x_i;z_i)||_{\Sigma_i}^2 \\
	\end{split}
\end{equation}

For the in-hand object tracking problem, we define states in the graph to be the 6-DOF object and end-effector poses at every time step $t=1\hdots T$, i.e.\ $x_t=[o_t\ e_t]^T$, where $o_t, e_t \in SE(3)$. Factors in the graph include image-to-image factors $f_{im2im}(\cdot)$, image-to-patch factors $f_{im2pc}(\cdot)$, velocity smoothness priors $f_{vel}(\cdot)$, end-effector pose priors $f_{eff}(\cdot)$ and vision priors for re-localization at the beginning of a contact episode $f_{vis}(\cdot)$. At every time step, new variables and factors are added to the graph. Writing out Eq.~\ref{eq:eq3.2} for our problem,
\begin{equation}
	\begin{split}
		\label{eq:eq3.3}
		\scriptsize
		& \hat{x}_{1:T} = \underset{x_{1:T}}{\operatorname{argmin}}\sum_{t=1}^{T} \left\{\right. ||f_{im2im}(o_{t-1}, o_{t}, e_{t-1}, e_{t})||^2_{\Sigma_{im2im}} + \\
		& ||f_{im2pc}(o_{t}, e_{t})||^2_{\Sigma_{im2pc}} + ||f_{vel}(o_{t-2}, o_{t-1}, o_{t})||^2_{\Sigma_{vel}} + \\
		& ||f_{eff}(e_{t})||^2_{\Sigma_{eff}} + ||f_{vis}(o_{t})||^2_{\Sigma_{vis}}\left.\right\}
	\end{split}
\end{equation}
Individual cost terms in Eq.~\ref{eq:eq3.3} are described in detail in Section \ref{subsec:approachGraphopt}. Eq.~\ref{eq:eq3.3} is the optimization objective that we must solve for every time step. Instead of resolving from scratch every time step, we make use of efficient, incremental solvers \cite{kaess2012isam2} to solve this in real-time.

\section{Approach}
\label{sec:approach}

We present a two-stage approach: First, we learn a mapping from tactile images to surface normals (Section \ref{subsec:approachLearnNormals}) which can be integrated to create a 3D reconstruction (Section \ref{subsec:approachReconNormals}). Second, we use these surface normals to create a 3D local patch map online within a factor graph for inferring the latent 3D object poses (Section \ref{subsec:approachGraphopt}).

\subsection{Learning surface normals}
\label{subsec:approachLearnNormals}

\begin{figure}[!t]
	\centering
	\includegraphics[width=0.9\columnwidth]{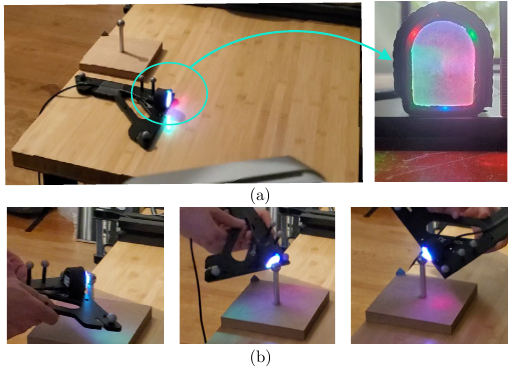}
	\caption{\footnotesize Real-world experimental setup showing: \textbf{(a)} a Sphere object mounted on a workbench with a close-up of the Digit sensor, and \textbf{(b)} a contact episode with the sensor moved around the object.}
	\label{fig:experimentalSetup}
	\figGap
\end{figure}

Here we discuss how we learn to predict surface normals using tactile color images from the Digit sensor \cite{lambeta2020digit}. Color images are generated by illuminating the gel surface with three light sources (Fig. \ref{fig:experimentalSetup}(a)), such that each light has a unique color and direction. When pressing an object into the gel, the gel surface conforms to the object. Under an idealized model of the sensor, the color and intensity of light reaching the camera due to diffused reflection from a point on the surface are directly related to the surface normal of the gel at that point. Hence, the RGB intensity values of image pixel are expected to contain significant information about the corresponding local surface normals of the object.

To infer surface normal images from tactile images we train an image translation network, pix2pix \cite{isola2017}. The pix2pix model is based on a generator-discriminator network architecture that enables it to learn mappings from a low amount of training data. It also enables us to learn a generalized mapping, i.e.\ we test on objects unseen during training. To train the model, we use a dataset of ground truth pairs of color and surface normal images $z=\{\mathcal{I}_{c}, \mathcal{I}_{n}\}$. We extract datasets in two ways. For simulation dataset, we generate surface normals by adding a normal shader to the Tacto simulator \cite{wang2020tacto} based on Pyrender \cite{matl2019pyrender}. For the real dataset, we follow a similar procedure as \cite{yuan2017gelsight, wang2021gelsight} of using a ball bearing of known radius whose ground truth normals can be synthesized. We manually annotate the circular patches in RGB images, and synthesize ground truth normals in the foreground annotations. We also augment our dataset with additional simulated images to improve generalization.

\subsection{Reconstruction from normals}
\label{subsec:approachReconNormals}

Given an image with local surface normal information, we generate a 3D representation of the corresponding surface geometry. This surface normal image $N(x,y)$ is related to the gradient of a corresponding depth image $z(x,y)$ as,
\begin{equation}
	\begin{split}
		\label{eq:eq4.2.1}
		\frac{\partial z}{\partial x}=\frac{n_x}{n_z},\hspace{20pt} \frac{\partial z}{\partial y}=\frac{n_y}{n_z}
	\end{split}
\end{equation}

Given depth gradients $\{\frac{\partial z}{\partial x}, \frac{\partial z}{\partial y}\}$, we can recover the depth map $z(x,y)$ by integration using a fast Poisson solver with discrete sine transform (DST) \cite{doerner2012poisson} as used in prior work \cite{yuan2017gelsight, wang2021gelsight}. For the boundary conditions, we use the mean distance to the undisturbed gel surface. A consequence of this choice is that contact regions crossing the edge of the image will not be correctly handled by the solver. We filter out such images in our trials.

Finally, once we have computed the depth map, we inverse project it to obtain a 3D point cloud. We use the OpenGL clip projection model \cite{openGLproj} to map from pixel to world coordinates, i.e.\ $\begin{bmatrix} x_{w}\ y_{w}\ z_{w} \end{bmatrix} = V P\begin{bmatrix} x_{pix}\ y_{pix}\ z_{depth} \end{bmatrix}$, where $P$ is the projection matrix based on near, far plane values of the projection frustum, and $V$ is the camera view matrix.



\subsection{Factor graph optimization}
\label{subsec:approachGraphopt}

Once we have the surface normal images, we integrate those along with other priors as factors within a factor graph. The factor graph optimizer then solves for the joint objective in Eq.~\ref{eq:eq3.3}. We look at each of the cost terms in Eq.~\ref{eq:eq3.3} in detail.

\subsubsection*{Image-to-image factors}
In some cases, it suffices to look at consecutive tactile images and infer the relative transformation between them. Color images from the tactile sensor at the current and previous time step are converted into point clouds $\{\mathcal{P}_{t-1}, \mathcal{P}_{t}\}$ in their respective end-effector or sensor frame. The two point clouds are registered against each other using a point-to-plane iterative closest point (ICP) algorithm. The resultant relative transformation is added as a binary factor between consecutive poses in the graph. This is expressed as the $f_{im2im}(.)$ term in Eq.~\ref{eq:eq3.3}, i.e.,
\begin{equation}
	\begin{split}
		\label{eq:eq4.c.1.1}
		||f_{im2im}(o_{t\text{-}1}, e_{t\text{-}1}, o_{t}, e_{t})||^2_{\Sigma_{im2im}}:=||{T}^{graph}_{t\text{-}1, t}\ominus T^{reg}_{t\text{-}1, t}||^{2}_{\Sigma_{im2im}}
	\end{split}
\end{equation}
where 
${T}^{graph}_{t\text{-}1, t}=(e_{t\text{-}1}o^{-1}_{t\text{-}1}) \ominus (e_{t}o_{t}^{-1})$ are current relative estimates from the graph and ${T}^{reg}_{t\text{-}1, t} = \argmin_{T} \sum_{i}||p^{(i)}_t - Tp^{(i)}_{t\text{-}1}||_2^2$ is the measured transformation from ICP registration. Both ${T}^{graph}_{t\text{-}1,t}$, ${T}^{reg}_{t\text{-}1, t}$ use gel center as their origin. $(p^{(i)}_{t-1}, p^{(i)}_t)\in (\mathcal{P}_{t-1}, \mathcal{P}_{t})$ are pairs of point correspondences in the two point clouds. $\ominus$ denotes difference between two SE(3) manifold elements.

\subsubsection*{Image-to-patch factors}
Image-to-image factors fail whenever the tactile image changes by a non-trivial amount leading to large registration errors. This happens whenever the object undergoes a larger transformation or moves in and out of the gel. To stabilize tracking in these situations we introduce a local patch model by fusing together multiple point clouds and register new tactile point cloud data against this patch. The local patch is maintained by fusing together images at specific key frames within the current contact episode, rather than fusing all available frames. We choose these key frames at fixed intervals given uniform motions but one can also select these based on a field-of-view overlap threshold. The current cloud $\mathcal{P}_{t}$ is registered against the local patch map cloud $\mathcal{P}_{map}$, and the relative transformation is added as factors to the graph,
\begin{equation}
	\begin{split}
		\label{eq:eq4.c.2.1}
		||f_{im2pc}(o_t, e_t)||^2_{\Sigma_{im2pc}}:=||{T}^{graph}_{map, t}\ominus {T}^{reg}_{map, t}||^{2}_{\Sigma_{im2pc}}
	\end{split}
\end{equation}
where ${T}^{graph}_{map, t}=o_{t}e_{t}^{-1}$ are current estimates from the graph and ${T}^{reg}_{map, t}= \argmin_{T} \sum_{i}||p^{(i)}_t - Tp^{(i)}_{map}||_2^2$ is the measured transformation from ICP registration. $(p^{(i)}_{t}, p^{(i)}_{map})\in (\mathcal{P}_{t}, \mathcal{P}_{map})$ are pairs of point correspondences in the current point cloud and local patch map.

\subsubsection*{Constant velocity priors}
We add a prior that assumes objects move at a constant velocity, which has the effect of smoothing tracked trajectories. This is a ternary factor between triplets of object poses,
\begin{equation}
	\begin{split}
		\label{eq:eq4.c.3.1}
		||f_{vel}(o_{t\text{-}2},o_{t\text{-}1}, o_{t})||^2_{\Sigma_{vel}}:=||o^{-1}_{t\text{-}2}o_{t\text{-}1}\ominus o^{-1}_{t\text{-}1}o_{t}||^{2}_{\Sigma_{vel}}
	\end{split}
\end{equation}

\subsubsection*{End-effector priors}
We model uncertainty about end-effector locations as unary priors on end-effector variables,
\begin{equation}
	\begin{split}
		\label{eq:eq4.b.3.1}
		||f_{eff}(e_{t})||^2_{\Sigma_{eff}}:=||e_t \ominus \tilde{e}^{mc}_t||^2_{\Sigma_{eff}} \\ 
	\end{split}
\end{equation}
where, $\tilde{e}^{mc}_t={e}^{mc}_t\oplus\ \mathcal{N}(0,\Sigma_{eff})$ are poses from the motion capture system with added Gaussian noise. We model the end-effectors as variables in the graph to keep the formulation general, especially when we have robot end-effector, the pose measurements would be coming from robot kinematics.

\subsubsection*{Vision prior}
We add a global vision pose prior for only the object pose at the start of an episode,
\begin{equation}
	\begin{split}
		\label{eq:eq4.c.4.1}
		||f_{vis}(o_{t})||^2_{\Sigma_{vis}}:=||o_t \ominus \tilde{o}^{vis}_t||^{2}_{\Sigma_{vis}}
	\end{split}
\end{equation}
where, $\tilde{o}^{vis}_t$ are poses with added Gaussian noise $\mathcal{N}(0,\Sigma_{vis})$. For re-localization during multi-contact episodes, we can add such a factor at the start pose of every new contact episode.

\section{Results and Evaluation}
We evaluate our approach qualitatively and quantitatively on a number of episodes where an object, unknown a priori, must be tracked from a sequence of tactile measurements. We compare against a set of baselines on two fronts: a) on surface normal predictions from images and b) on the final tracking error of object poses. We use PyTorch \cite{paszke2019pytorch} for training surface normal models, and GTSAM C++ library \cite{dellaert2012factor} for factor graph optimization. We specifically use the iSAM2 \cite{kaess2012isam2} solver for efficient, real-time optimization.

\begin{figure}[!b]
	\centering
	\includegraphics[width=\columnwidth]{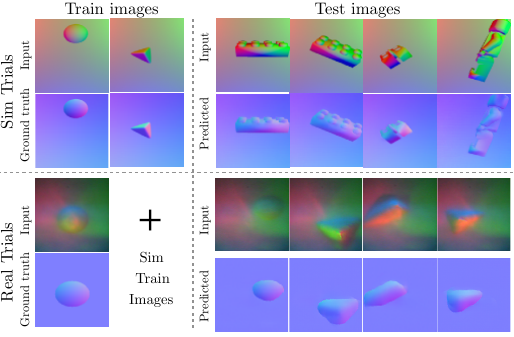}
	\caption{\footnotesize Training and evaluation of the pix2pix (unet) model that maps input color images to predicted surface normal images.}
	\label{fig:resultsSurfaceNormals}
	\vspace{-3.5mm}
\end{figure}

\begin{table}[!b]
	\centering
 	\caption{\small Learned surface normal performance (validation loss)}
	\label{tab:surfaceNormalModelPerf}
	\setlength{\tabcolsep}{4pt}
	\resizebox{\columnwidth}{!}{
	\begin{tabulary}{\textwidth}{LCCCCC} 
		\multirow{3}{*}{\textbf{Model type}} & & & \textbf{Object Shapes} & \\ \cmidrule{2-6}
		& \textbf{Sim sphere} & \textbf{Cube} & \textbf{Real sphere} & \textbf{Toy brick} & \textbf{Toy human} \\ \midrule
		{pix2pix (unet)} & \textbf{0.4e-3} & \textbf{0.5e-3} & \textbf{1.0e-3} & \textbf{1.1e-3} & \textbf{0.9e-3} \\
		{pix2pix (resnet)} & 0.6e-3 & 1.2e-3 & 1.6e-3 & \textbf{1.1e-3} & 1.0e-3 \\
		{MLP 3-layer} & 1.4e-3 & \textbf{0.5e-3} & 5.8e-3 & 5.1e-3 & 6.7e-3 \\ \midrule
	\end{tabulary}
	}
\end{table}

\begin{figure*}[!t]
	\centering
	\includegraphics[width=0.87\textwidth]{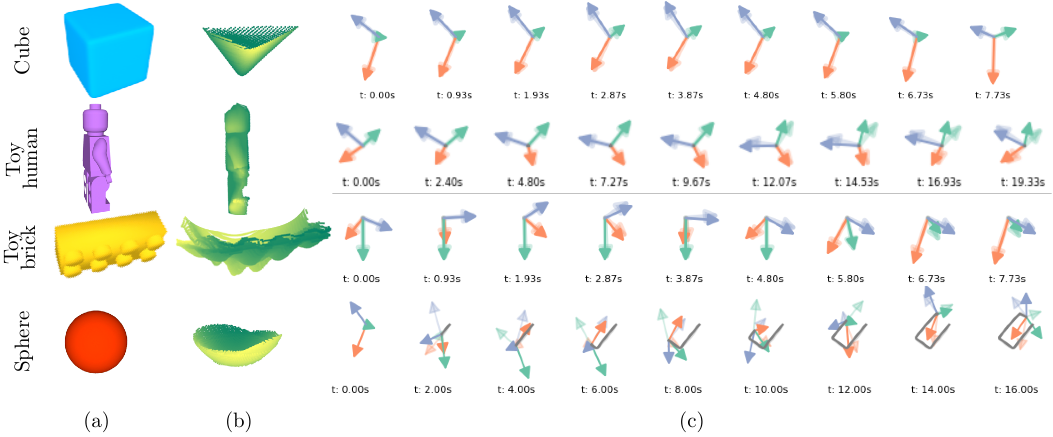}
	\caption{\footnotesize Qualitative pose tracking performance on simulated trials. \textbf{(a)} Object being tracked. \textbf{(b)} Local patch map reconstructed online. \textbf{(c)} Estimated 3D pose coordinates  (light) against ground truth (dark). Rotations in orange (x), purple (y), green (z). Translations in grey.\figGap}
	\label{fig:resultsSimQual}
\end{figure*}

\begin{figure*}[!t]
	\centering
	\includegraphics[width=0.87\textwidth]{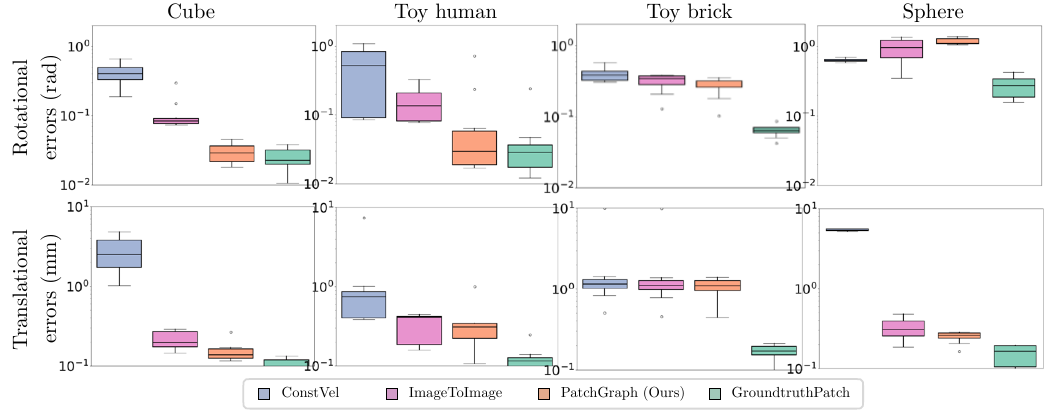}
	\caption{\footnotesize Final tracking error box plots on simulated trials: \textbf{(top)} rotation and \textbf{(bottom)} translation errors in semi-log scale.}
	\label{fig:resultsSimQuant}
	\figGap
\end{figure*}

\subsection{Experimental setup}


\subsubsection*{Simulator}
We collected simulation data using the Tacto~\cite{wang2020tacto} simulator where one can load the Digit sensor, an object and render high-resolution tactile image readings in real-time. The simulator uses PyBullet \cite{coumans2016pybullet} as the underlying physics engine and Pyrender \cite{matl2019pyrender} as the back-end rendering engine for generating images. We generated trials by using a position controller to move the object on the sensor surface. We collect data for a diverse set of objects: Sphere, Cube, Toy human and Toy brick.

\subsubsection*{Real-world}
For real-world episode, we used a Digit \cite{lambeta2020digit} sensor to get tactile measurements. We mounted the object on a workbench and mounted the Digit on a movable plate, both of which were tracked by an OptiTrack motion capture system to get ground truth poses. We collect real-world data for two objects: a Sphere (ball-bearing) of 1/2" diameter and a Pyramid 1/2" tall with 1.75" side length. 

\begin{figure*}[!t]
	\centering
	\includegraphics[width=0.88\textwidth]{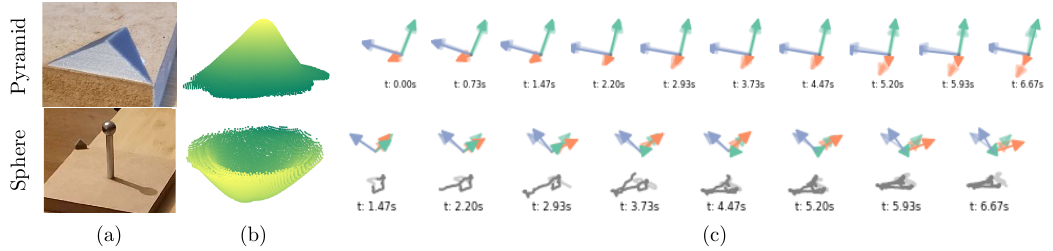}
	\caption{\footnotesize Qualitative pose tracking performance on real trials. \textbf{(a)} Object being tracked. \textbf{(b)} Local patch map reconstructed online. \textbf{(c)} Estimated 3D pose coordinates  (light) against ground truth (dark). Rotations in orange (x), purple (y), green (z). Translations in grey.}
	\label{fig:resultsRealQual}
	\figGap
\end{figure*}

\begin{figure}[!t]
	\centering
	\includegraphics[width=0.85\columnwidth]{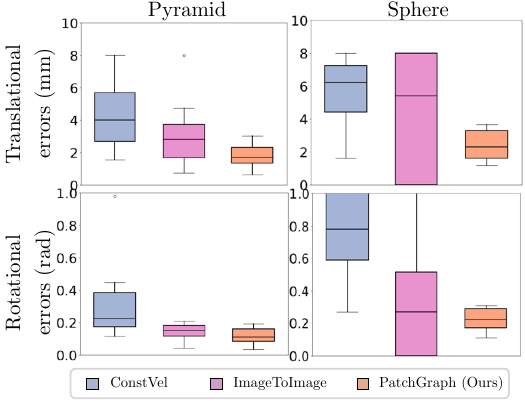}
	\caption{\footnotesize Final tracking error box plots on real trials: \textbf{(top)} rotation and \textbf{(bottom)} translation errors.}
	\label{fig:resultsRealQuant}
	\figGap
\end{figure}

\subsection{Surface normal reconstructions}
We first analyze the accuracy of learned surface normals from color images collected by the Digit tactile sensor. For simulation episodes, we only train on two simple objects: 50 images of Sphere and 50 images of Cube. We tested the model on two different held-out shapes: Toy brick and Toy human. For real episodes, we only had ground truth normals for the real sphere. Hence, we expanded the training dataset to include both the training data from simulation and the real Sphere data. We tested the model on held-out real Pyramid object. For baselines, we picked two different pix2pix architectures, unet and resnet. We also trained a baseline 3-layer MLP 5-32-32-3 with tanh activation on an L2 loss, mapping a single color pixel (r,g,b,x,y) to a surface normal (nx,ny,nz), similar to prior work \cite{wang2021gelsight}.

\subsubsection*{Qualitative reconstructions} Fig. \ref{fig:resultsSurfaceNormals} shows qualitative performance of the pix2pix (unet) model on both real and simulated images. We can see the model generalizes to fairly different shapes such as the Toy human and Toy brick, even though it is trained on simple objects. Moreover, fine-tuning this model with very little real data, i.e., 50 images of the real Sphere, enables it to generalize to unseen geometries in the real-world such as different local patches of a Pyramid.

\subsubsection*{Quantitative model performance} Table \ref{tab:surfaceNormalModelPerf} compares mean-squared pixel loss on the validation dataset for different model choices. We see that pix2pix, for both unet and resnet architecture, has a fairly low MSE loss and generalizes to unseen shapes such as the Toy human and Toy brick. On the other hand, the per-pixel MLP baseline incurs a high MSE loss. In general, the per-pixel MLP is likely to be insufficient for non-ideal tactile sensors with effects such as self-shadowing of the gel. Having convolutional layers, such as with the pix2pix architectures, can address such confounding effects since these are typically localized in the image, e.g.\ shadows are cast by nearby object features since object depth is small relative to the image size.


\subsection{Factor graph optimization}
We now look at the final task performance of tracking 3D object poses using tactile image measurements. We compare 4 objectives, each of which uses different factors: \textit{ConstVel} uses only a constant velocity prior, \textit{ImageToImage} uses image-to-image factors, \textit{PatchGraph} (ours) additionally uses Image-to-patch factors (Fig. \ref{fig:approachOverall}) and \textit{GroundtruthPatch} uses a global object model. \textit{GroundtruthPatch} assumes the object is known a-priori and hence represents the best a method can do. We keep the covariance parameters same across all runs in the graph, i.e. $\Sigma_{im2im}\text{=}1e\text{-}3\times\mathbb{I}_{6\times6}$, $\Sigma_{im2pc}\text{=}1e\text{-}3\times\mathbb{I}_{6\times6}$, $\Sigma_{vel}\text{=}1e\text{-}2\times\mathbb{I}_{6\times6}$, $\Sigma_{eff}\text{=}1e\text{-}5\times\mathbb{I}_{6\times6}$. First pose of the object is assumed to be known, and added as a unary prior to graph.

\subsubsection*{Simulation tracking performance}
Fig. \ref{fig:resultsSimQual} shows the qualitative tracking performance of \textit{PatchGraph} for various objects in simulated trials. 
For Cube, Toy brick and Toy human, \textit{PatchGraph} is able to reliably track rotations of the object. The local patch, constructed from online estimates, appears to be consistent with the local object geometry, thus explaining the good tracking performance. The object that was most difficult to track was Toy brick, as evidenced by the distortions in the patch, owing to jerkiness of the in-contact motions. We also note that tracking rotations for a Sphere has high errors, which is expected since rotations of a sphere are unobservable from tactile images.

Fig. \ref{fig:resultsSimQuant} shows the quantitative rotation and translation tracking performance against baselines on all objects, with $20$ distinct contact sequences per object. Overall, \textit{PatchGraph} has the lowest errors matching closely to that of \textit{GroundtruthPatch} that has the global object model. This lends credence to our claim that a local patch suffices for reliable object tracking. \textit{ConstVel} understandably has the highest variance among any baselines. \textit{ImageToImage} fails to outperform \textit{PatchGraph} on any of the datasets. Toy brick appears to be the most challenging among datasets, where \textit{GroundtruthPatch} has a clear performance gap. 
\subsubsection*{Real tracking performance}
Fig.~\ref{fig:resultsRealQual} shows qualitative tracking performance of \textit{PatchGraph} for various objects in real trials. Fig. \ref{fig:resultsRealQual}(b) shows the local patches which appear consistent with the local object geometry, and is a key piece to reliable tracking. Fig. \ref{fig:resultsRealQual}(c) shows good rotation tracking for Pyramid and translation tracking for Sphere.

Fig.~\ref{fig:resultsRealQuant} shows quantitative rotation and translation tracking performance against baselines on all objects, with $10$ distinct contact sequences per object. The main observation is that \textit{ImageToImage} performs much worse than \textit{ImageToPatch}, particularly in translation errors for Sphere. This is primarily because point clouds generated from individual images are not as geometrically discriminative as the fused local patch, causing \textit{ImageToImage} factors alone to diverge quickly. \textit{PatchGraph} keeps translation errors under 4mm and rotation errors under 0.2rad, which looks promising for use in dexterous object manipulations.

\section{Conclusion}
We presented a factor graph-based approach for tracking 3D object poses from tactile image sequences during in-hand manipulations. We showed reliable tracking on $4$ simulated objects and $2$ real objects without relying on any a priori object information. We achieved this by exploiting two decompositions of the tracking problem. First, that a complex object can be treated as a composition of many local patches each of which can be mapped and tracked largely independently. Second, surface normal information is highly localized within a tactile image and independent of the global object shape.

A primary limitation that can cause tracking failures is when the local patch map is not sufficiently discriminative geometrically, e.g. flat or featureless patches, or when the patch motions are degenerate in the observed image space, e.g. rotations of a spherical object. As future work, it would be interesting to explore solutions that take into account geometric degeneracies \cite{zhang2016degeneracy, westman2019degeneracy} as well as approaches that can detect slip and shear \cite{yuan2015measurement} to disambiguate motion degeneracies. Another interesting future direction would be to complement the tracker with a global first pose re-localization that is able to generalize across objects, e.g. using visual images to predict a contact location likelihood.


\section*{\footnotesize{Acknowledgements}}
\footnotesize{This work was supported through the FRAIM program. We would like to thank Wenzhen Yuan for insightful feedback and suggestions. We also thank the DIGIT team, particularly Roberto Calandra and Mike Lambeta, for helpful discussions and support with the sensor and software.}

\balance




\footnotesize
\bibliographystyle{ieeetr}
\clearpage
\bibliography{references}

\end{document}